\documentclass[letterpaper, 10 pt, conference]{ieeeconf}  

\IEEEoverridecommandlockouts                              

\usepackage{graphics} 
\usepackage{epsfig} 
\usepackage{mathptmx} 
\usepackage{times} 
\usepackage{amsmath} 
\usepackage{amssymb}  
\usepackage{booktabs}
\usepackage{threeparttable} 
\usepackage{graphicx}
\usepackage{algorithm}
\usepackage{algpseudocode}
\usepackage{subfig}
\usepackage{cite}
\usepackage{xcolor}
\usepackage{subcaption}
\usepackage{hyperref}

\title{\LARGE \bf
AnywhereVLA: Language-Conditioned Exploration and Mobile Manipulation
}

\author{
    \parbox{14cm}{\centering
        Konstantin Gubernatorov*, Artem Voronov*, Roman Voronov*, Sergei Pasynkov*,\\
        Stepan Perminov, Ziang Guo, and Dzmitry Tsetserukou
    }
    \thanks{*Denotes equal contribution.}
    \thanks{All authors are with the Intelligent Space Robotics Laboratory, Center for Digital Engineering, Skolkovo Institute of Science and Technology, Moscow, Russia. \tt \{{Konstantin.Gubernatorov}, {Artem.Voronov}, {Roman.Voronov}, {Sergei.Pasynkov}, {Stepan.Perminov}, {Ziang.Guo}, {D.Tsetserukou}\} @skoltech.ru}
}

\begin{document}

\maketitle
\thispagestyle{empty}
\pagestyle{empty}

\begin{figure*}[t]
    \centering
    \includegraphics[width=\textwidth]{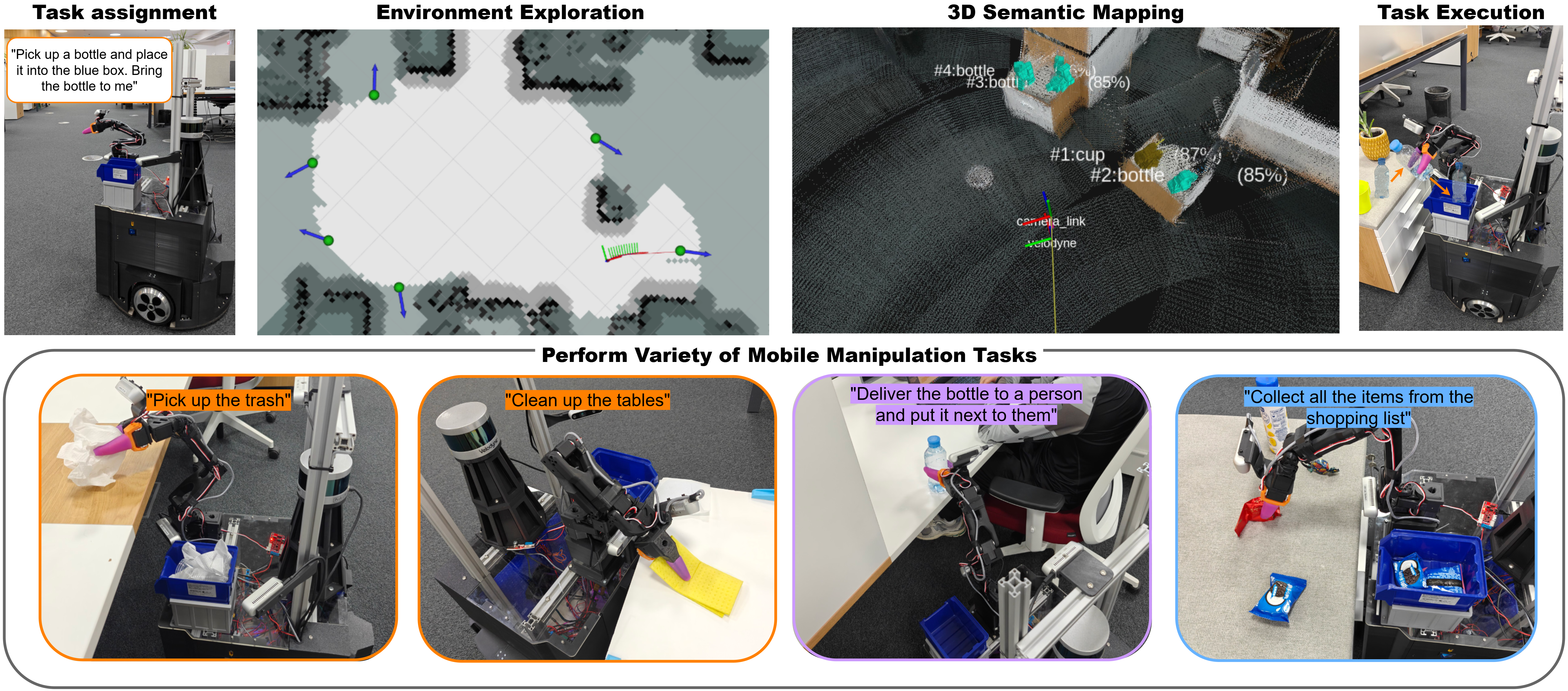}
    \caption{AnywhereVLA is the modular architecture comprising VLA manipulation and environment exploration. Given the task, AnywhereVLA parses it into simpler actions which further condition Active Environment Exploration. Exploration and navigation in larger-scale indoor environments are performed within a 3D point cloud semantic map. By leveraging a purpose-built pick-and-place dataset, AnywhereVLA exhibits robust generalization capacities.}
    \label{fig:teaser}
\end{figure*}

\begin{abstract}
We address natural language pick-and-place in unseen, unpredictable indoor environments with AnywhereVLA, a modular framework for mobile manipulation. A user text prompt serves as an entry point and is parsed into a structured task graph that conditions classical SLAM with LiDAR and cameras, metric semantic mapping, and a task-aware frontier exploration policy. An approach planner then selects visibility and reachability aware pre grasp base poses. For interaction, a compact SmolVLA manipulation head is fine tuned on platform pick and place trajectories for the SO-101 by TheRobotStudio, grounding local visual context and sub-goals into grasp and place proposals. The full system runs fully onboard on consumer-level hardware, with Jetson Orin NX for perception and VLA and an Intel NUC for SLAM, exploration, and control, sustaining real-time operation. We evaluated AnywhereVLA in a multi-room lab under static scenes and normal human motion. In this setting, the system achieves a $46\%$ overall task success rate while maintaining throughput on embedded compute. By combining a classical stack with a fine-tuned VLA manipulation, the system inherits the reliability of geometry-based navigation with the agility and task generalization of language-conditioned manipulation. All code, models, and datasets are open source and are available on the \href{https://selfai-research.github.io/AnywhereVLA/}{\textcolor{blue}{project GitHub repository}}.
\end{abstract}

\section{Introduction}

Mobile manipulation is accelerating beyond limited indoor workcells towards large unstructured environments, in which robots need to explore unfamiliar cluttered spaces and physically interact with diverse objects and people. The execution of complex mobile manipulation tasks conditional on natural language instructions has gained attention in recent years in the field of service robotics\cite{thakar2023survey}. Research on intelligent robotic systems in fields such as household service \cite{wu2024tidybotopensourceholonomicmobile, fu2024mobilealohalearningbimanual}, retail automation \cite{huang2025practicalinsightsgraspstrategies, bajracharya2024demonstrating}, warehouse logistics \cite{he2023tacmms}, and manufacturing \cite{pu2023general} has gained popularity, highlighting the importance of developing mobile manipulation solutions capable of operating in large-scale and open-plan indoor environments. Recent studies have increasingly focused on natural language processing to enable robots to interpret human instructions and facilitate intuitive task specification \cite{Park_2024}, positioning language-guided manipulation as a key approach for effective human-robot collaboration \cite{zhou2025languageguidedlonghorizonmanipulation}. However, unifying language-based control, environment exploration, and manipulation in expansive environments presents a significant challenge \cite{shao2025largevlmbasedvisionlanguageactionmodels}.

Vision-language-action (VLA) models show strong generalization in various mobile manipulation tasks \cite{black2024pi0visionlanguageactionflowmodel, intelligence2025pi05visionlanguageactionmodelopenworld}, enabling robots to perform complex operations integrating perception, language, and control. Despite these advancements, several critical limitations persist for end-to-end control use for mobile robots. Most VLA models are confined to specific tasks and have limited spatial awareness \cite{black2024pi0visionlanguageactionflowmodel, intelligence2025pi05visionlanguageactionmodelopenworld, wu2025momanipvlatransferringvisionlanguageactionmodels}, restricting their operational scope to localized settings and impeding their ability to navigate or manipulate objects in unseen or occluded regions of larger indoor spaces. 

Vision-Language Navigation (VLN) approaches \cite{shah2024bumbleunifyingreasoningacting} present an end-to-end VLM-based framework that navigates building-wide environments while manipulating previously unseen household objects. However, \cite{shah2024bumbleunifyingreasoningacting}, as all VLN models, require instructions about the location of the target object within the environment, which is often impractical in dynamic or unexplored settings. In contrast, classical navigation stacks \cite{macenski2020marathon2} provide robust solutions for mapping and environment exploration, enabling robots to systematically traverse and model unknown spaces. However, these traditional systems lack the advanced language comprehension and semantic reasoning capabilities necessary to interpret complex instructions or contextual cues \cite{Park_2024}, limiting their ability to perform goal-directed tasks that require understanding natural language or high-level semantic objectives.

In this paper we introduce AnywhereVLA, a novel modular architecture for large-scale indoor mobile manipulation, addressing the limitations of existing Vision-Language-Action (VLA) models that adapt pretrained vision-language models (VLMs) to enable natural language-driven perception and control, but are often constrained to room-scale environments due to high computational demands. Drawing from advancements in VLA paradigms, AnywhereVLA integrates the rich visual and linguistic knowledge encoded in VLMs—leveraged through co-fine-tuning on robotic trajectory data and Internet-scale vision-language tasks—with the robust traversability afforded by classical navigation stacks, representing robot actions as tokenized sequences to facilitate end-to-end control and emergent semantic reasoning.

AnywhereVLA is a pipeline for large-scale indoor mobile manipulation in unseen environments. As shown in Fig.~\ref{fig:teaser}, AnywhereVLA combines robust traversability of classical navigation algorithms and simultaneous localization and mapping (SLAM) with the generalizable scene understanding and task grounding of VLA models. Our pipeline translates high-level language instructions into low-level control commands by generating actions via VLA model for task-specific manipulation and computes navigation trajectories via language-conditioned exploration algorithm, directly actuating the wheels of the mobile base and the joints of the manipulator. Our main contribution is the following:

We propose a cohesive modular framework that accepts a single language-based task instruction as input, which conditions the environment exploration and navigation modules and simultaneously drives the VLA model for manipulation task execution. Our system achieves real-time performance exceeding 10 Hz across all modules that are deployed on consumer-available edge computing units, ensuring efficient and responsive operation in dynamic settings.

\begin{figure*}[t]
    \centering
    \includegraphics[width=\textwidth]{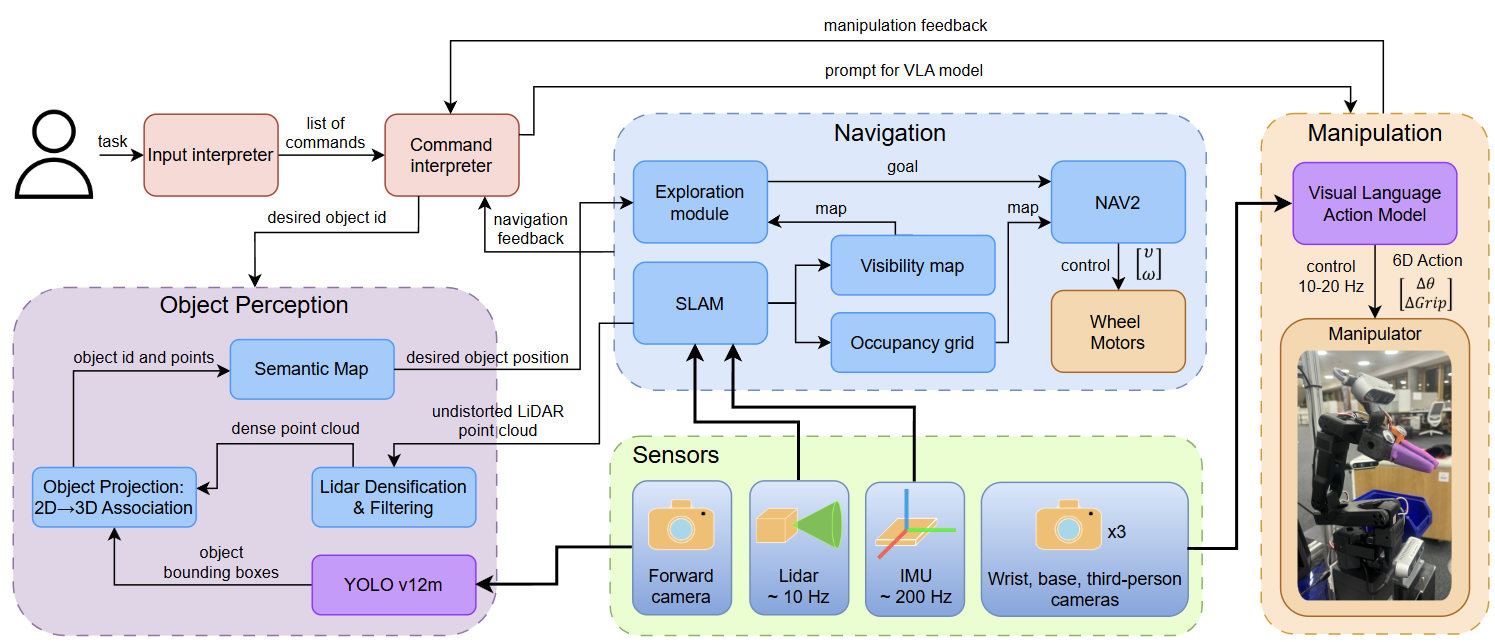}
    \caption{AnywhereVLA architecture.}
    \label{fig:architecture}
\end{figure*}

\section{Related Work}

Large Vision-Language Models (VLMs) have emerged as a transformative paradigm. Pretraining on web-scale image-text corpora enables robust alignment between visual content and natural language semantics. When incorporated into robotic systems, these models have catalyzed a new class of policies, Vision-Language-Action (VLA) models, which extend perception-language alignment to action generation by mapping open instructions and visual context to executable behaviors \cite{kim2024openvlaopensourcevisionlanguageactionmodel}. VLA models unify perception, natural language, and control into end-to-end policies for robots. They leverage pretrained vision-language models and robotic trajectory datasets to ground instructions into low-level actions. MoManipVLA \cite{wu2025momanipvlatransferringvisionlanguageactionmodels} adapts pretrained VLA models to mobile manipulators by jointly planning base and arm motions, enabling complex household operations. $\pi_0$ \cite{black2024pi0visionlanguageactionflowmodel} introduces a flow matching-based action generation approach that integrates Internet-scale semantic knowledge and demonstration data for zero-shot dexterous manipulation, while $\pi_{0.5}$ \cite{intelligence2025pi05visionlanguageactionmodelopenworld} extends this paradigm through co-training on heterogeneous multimodal datasets, achieving strong generalization across tasks and robots. These methods excel at instruction grounding and task generalization, but have limited spatial awareness.

Beyond this, the deployment of resource-intensive VLAs on mobile platforms requires careful optimization to ensure efficiency and maintain performance within hardware limitations. SmolVLA \cite{shukor2025smolvlavisionlanguageactionmodelaffordable} demonstrates that a 450M-parameter VLA can achieve competitive performance compared to larger models. EdgeVLA \cite{budzianowski2025edgevlaefficientvisionlanguageactionmodels} further improves efficiency by eliminating autoregressive decoding for end-effector prediction, yielding a 7$\times$ speedup in inference while maintaining task accuracy. TinyVLA \cite{wen2025tinyvlafastdataefficientvisionlanguageaction}  introduces a diffusion-based policy decoder and a lightweight multimodal backbone, that match the performance of much larger VLAs while being significantly faster and more data-efficient. These advances enable real-time operation on embedded devices, though the generalization reported is primarily within manipulation domains.

A rapidly expanding body of work explores diffusion transformer-based policies for language-conditioned control. RDT-1B \cite{liu2025rdt1bdiffusionfoundationmodel} is a billion-parameter diffusion policy pretrained on multi-robot datasets, enabling broad semantic generalization and robust bimanual skills. AC-DiT \cite{chen2025acditadaptivecoordinationdiffusion} introduces an adaptive coordination transformer that conditions manipulation policies in the mobility context, facilitating coupled base and arm control in mobile manipulators. However, both \cite{liu2025rdt1bdiffusionfoundationmodel} and \cite{chen2025acditadaptivecoordinationdiffusion} focus exclusively on manipulation and therefore require complementary modules to provide spatial memory, target discovery, and long-horizon navigation.

BUMBLE \cite{shah2024bumbleunifyingreasoningacting} introduces a VLM-based end-to-end framework for building-wide mobile manipulation. As a fully integrated system, it demonstrates strong performance in navigating large-scale indoor environments while manipulating a broad and previously unseen set of everyday objects. A central limitation, however, is its reliance on the known map of the whole environment and set of pre-provided landmarks. In contrast, approaches that integrate SLAM and environment exploration overcome this limitation by navigating environments autonomously without requiring prior knowledge.

Furthermore, ASC \cite{yokoyama2023ascadaptiveskillcoordination} frames long-horizon mobile manipulation as a sequence of modular visuomotor skills coordinated by a high-level policy. Experiments in a $185\,\mathrm{m}^2$ apartment show reliable pick-and-place with strong spatial memory, indicating effective maintenance of the task context in the rooms and for extended periods. However, complete exploration and manipulation in apartment-sized environments takes 10-15 minutes. Consequently, while ASC substantiates scalable skill coordination, its motion efficiency remains a bottleneck for deployment in real-world, human-populated spaces.

\section{anywhereVLA Architecture}
AnywhereVLA framework implements an modular pipeline actuating motors of the mobile platform and manipulator's joints processing single language command and raw sensor inputs. In Fig. \ref{fig:architecture}, the architecture comprises four main modules: 3D Semantic Mapping with Confidence (SM), Active Environment Exploration (AEE), Approach, and VLA Manipulation. The workflow begins with the parsing of natural language instruction, which simultaneously informs the VLA module for task-specific manipulation and conditions the AEE process. The SM module constructs a semantic 3D point cloud map by combining LiDAR-Inertial-Visual SLAM with semantic annotations from the object detection model. This map supports the AEE module, which employs frontier-based exploration conditioned on the target object class derived from the language instruction. Exploration halts once the target object is detected and localized within the semantic map. The Approach module then navigates the mobile base to the target object's location using a 2D grid map projected from a filtered LiDAR point cloud, positioning the robot for manipulation. VLA Manipulation module, leveraging a fine-tuned SmolVLA model, executes the specified task by generating actions for the manipulator's motors.

\subsection{3D Semantic Mapping with Confidence (SM)}
Algorithm \ref{alg:semantic_map} presents a 3D Semantic Object Map construction by synchronizing RGB images, undistorted LiDAR point clouds, and 2D boundbox detections by projecting LiDAR points into the camera frame to associate them with detections. To mitigate the sparsity and discontinuities of spinning LiDAR, we densified the scan by interpolating between adjacent elevation rings within each azimuth bin, followed by voxelization. For each detection, near-surface points inside its enlarged 2D bounding box are backprojected to 3D, forming object-wise point sets that are transformed to the world frame and accumulated per class over time. This produces a metric-semantic map populated by per-object 3D statistics and confidence estimates.

\subsection{Object Aggregation}
Per class accumulated points are clustered via radius-based DBSCAN \cite{Ester1996ADA}, outliers are robustly filtered, and each cluster is summarized by its centroid and covariance. Multi-view and data-driven cues produce an object-level confidence via a logistic mapping that fuses point density, angular coverage, inlier count, and detector scores. The resulting world-frame semantic object map exposes persistent objects with class labels, geometry, and confidence for downstream planning and manipulation.

\begin{algorithm}[h]
\caption{Semantic Map Construction Pipeline}
\label{alg:semantic_map}
\begin{algorithmic}[1]
\State \textbf{Inputs:} image $I_t$, detections $\mathcal{D}_t$, LiDAR cloud $\mathcal{P}^L_t$, extrinsics $\mathbf{T}_{L\rightarrow C}$, TF $\mathbf{T}_{L\rightarrow W}$
\State \textbf{FOV filter:} project $\mathcal{P}^L_t$ to camera: $\mathcal{U}_t \leftarrow \Pi\!\left(\mathbf{T}_{L\rightarrow C}\,\mathcal{P}^L_t\right)$; keep points inside image  \\
\textbf{Densification:}
\For{each azimuth bin $k$}
\For{each adjacent ring pair $(r,r{+}1)$ with points $\mathbf{S}_k^r,\mathbf{E}_k^{r+1}$ meeting range/gap gates}
  \State insert $M$ interpolants $\mathbf{P}_t = \frac{M+1-t}{M+1}\mathbf{S}_k^r + \frac{t}{M+1}\mathbf{E}_k^{r+1}$, $t{=}1{:}M$
\EndFor
\EndFor
\State \textbf{Voxelize:} $\mathcal{Q}_t \leftarrow VoxelGrid(\text{FOV points},\, v)$
\\
\textbf{Projection:}
\For{each detection $b \in \mathcal{D}_t$}
  \State enlarge box $(w,h)\!\leftarrow\!(w{+}\Delta_x,h{+}\Delta_y)$, select points in 2D box
  \State depth gate: keep points with $z \le z_{\min}(b) + \delta$
  \State compute 3D bbox $(\mathbf{c},\mathbf{s})$ from selected points in LiDAR frame
  \State transform points to world: $\mathcal{Q}^W_b \leftarrow \mathbf{T}_{L\rightarrow W}(t)\,\mathcal{Q}_b$
  \State accumulate $\mathcal{Q}^W_b$ into per-class database
\EndFor
\\
\textbf{Object aggregation:}
\For{each class $\kappa$ (periodic/on-demand)}
  \State aggregate points; cluster with DBSCAN$(\varepsilon,\text{minPts})$
  \For{each cluster $C$}
    \State robust outlier filtering (MAD/$\sigma$); compute mean $\boldsymbol{\mu}$ and covariance $\boldsymbol{\Sigma}$
    \State estimate confidence $C$ using Eq.~\eqref{eq:confidence}; publish markers and metadata
  \EndFor
\EndFor
\end{algorithmic}
\end{algorithm}

\subsection{LiDAR densification}
\begin{figure}[t]
    \centering
    \includegraphics[width=\linewidth]{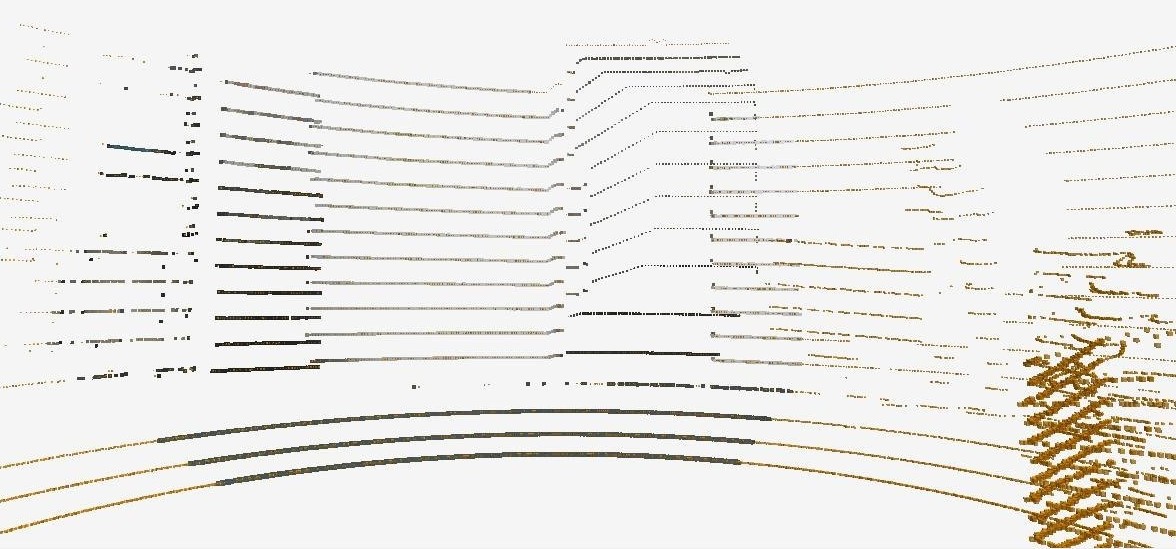}
    \caption{Sparse point cloud from Velodyne VLP-16 LiDAR.}
    \label{fig:lidar_init}
\end{figure}

\begin{figure}[H]
    \centering
    \includegraphics[width=\linewidth]{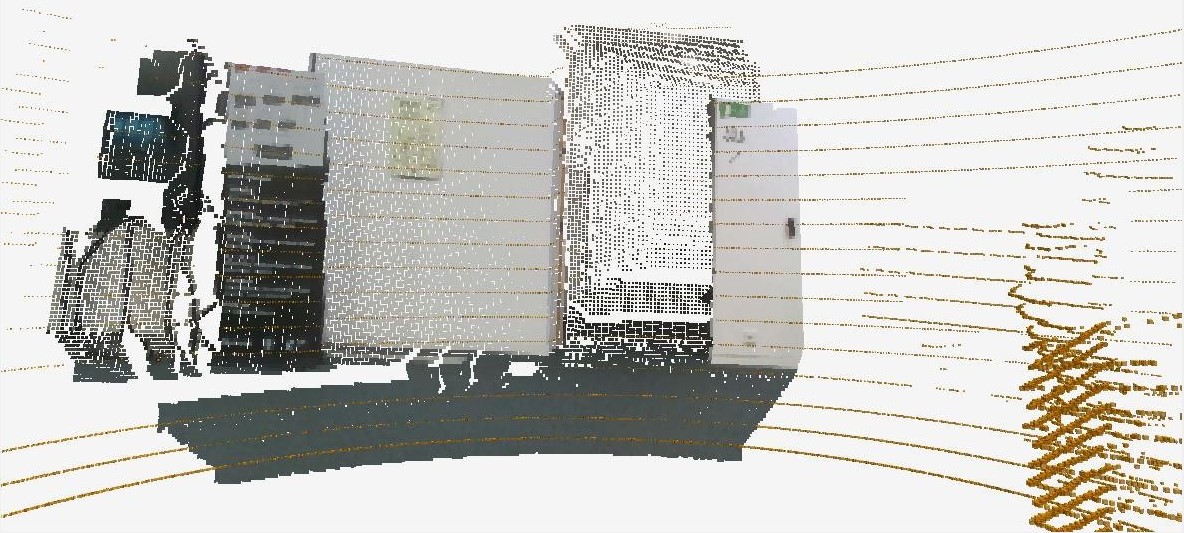}
    \caption{Densified LiDAR point cloud after interpolation.}
    \label{fig:lidar_dense}
\end{figure}

In Fig. \ref{fig:lidar_init}, the sparse and discontinuous LiDAR scan pattern causes few (or no) valid returns inside a 2D detection box. In Fig. \ref{fig:lidar_dense}, we interpolate between valid ring-adjacent samples within the same azimuth bin, inserting $M$ interior points per gap (subject to range-jump and spatial-gap gates). With endpoints $\mathbf{S}$ and $\mathbf{E}$, the interpolants are
\begin{equation}
\mathbf{P}_t \;=\; \frac{M+1-t}{\,M+1\,}\,\mathbf{S} \;+\; \frac{t}{\,M+1\,}\,\mathbf{E}, \qquad t=1,\ldots,M~.
\label{eq:densify}
\end{equation}

\subsection{Confidence estimation}
Let $\rho$ denote point density (normalized by $\rho_0$), $\Omega\!\in[0,1]$ denote multi-view angular coverage (union of yaw arcs over $2\pi$), $N$ the inlier count (normalized by $N_0$) and $\bar{s}$ the mean detector score. The confidence is
\begin{equation}
\begin{split}
C \;=\; \sigma\!\Big( \;w_\rho\big(1-e^{-\rho/\rho_0}\big)\;+\;w_\Omega\,\Omega\; \\ +\;w_N\big(1-e^{-N/N_0}\big)\;+\;w_S\,\bar{s}\;+\;b\;\Big)\,,
\end{split}
\label{eq:confidence}
\end{equation}
where $\sigma(\cdot)$ is the logistic function and $w_\rho,w_\Omega,w_N,w_S,b$ are tunable parameters.

\subsection{Active Environment Exploration (AEE)}

In Algorithm \ref{alg:goal_selection}, AEE module systematically maps unknown regions to locate the target object using a frontier-based strategy~\cite{yamauchi1997frontier} integrated into our modular pipeline. The occupancy grids of the SLAM backend serve as input. Frontiers are extracted through 8-connected morphological dilation and clustered. For each cluster, we compute a centroid and optimize the yaw to improve field-of-view (FoV) coverage. The resulting goal is validated with the Nav2 stack~\cite{macenski2020marathon2}. To suppress spurious goals and respect operational limits, we apply cluster filtering \((\eta_c=20\,\text{px})\), chunking \((\eta_k=50\,\text{px})\), non-maximum suppression \((d_{\min}=1\,\mathrm{m})\), an exploration radius constraint $R_e$, and FoV yaw optimization \((\alpha=35^{\circ},\,R_g=1.5\,\mathrm{m})\). We also use a conservative frontier filter with a \(5\times5\) kernel and a gain threshold of 50. The planner re-evaluates goals every \(T_u=4\,\mathrm{s}\). Exploration ends upon detection of the target or upon exhaustion of valid frontiers.

\begin{algorithm}
\caption{Frontier-Based Goal Selection}
\label{alg:goal_selection}
\begin{algorithmic}[1]
\Require Map $\mathcal{M}$, robot pose $(\mathbf{p}_r, \psi_r)$, camera params $(\alpha, R_g)$
\Ensure Goal pose $\mathbf{g}$ or $\bot$ (none)

\State $\mathbf{F} \gets$ Dilate(free($\mathcal{M}$), $3\times3$) $\cap$ unknown($\mathcal{M}$) $\setminus$ Dilate(occupied($\mathcal{M}$), $5\times5$)
\State Labels, Slices $\gets$ LabelComponents($\mathbf{F}$)

\For{each slice $s_i$ in Slices}
    \State Pixels $\gets$ Extract($s_i$)
    \If{$|\text{Pixels}| < \eta_c$}
        \State \textbf{continue}
    \EndIf
    \State Chunks $\gets$ AngularPartition(Pixels, $\eta_k$)
    \For{each chunk $c_j$ in Chunks}
        \State $\mathbf{p}_{ij} \gets$ Centroid($c_j$) $\oplus$ WorldTransform
        \State $\mathbf{p}_{ij} \gets$ Standoff($\mathbf{p}_{ij}$, $\mathbf{p}_r$, $d_s$)
        \If{$\|\mathbf{p}_{ij} - \mathbf{p}_r\| > R_e$}
            \State \textbf{continue}
        \EndIf
        \State $\psi_{ij} \gets$ OptimizeYaw($\mathbf{p}_{ij}$, unknown($\mathcal{M}$), $\alpha$, $R_g$)
        \If{Gain($\psi_{ij}$) $< 50$}
            \State \textbf{continue}
        \EndIf
        \State Add $(\mathbf{p}_{ij}, \psi_{ij})$ to Candidates
    \EndFor
\EndFor

\State Candidates $\gets$ NMS(Candidates, $d_{\min}$); Sort by $\|\mathbf{p} - \mathbf{p}_r\|$
\For{each $(\mathbf{p}_i, \psi_i)$ in Candidates}
    \State feasible, len $\gets$ ComputePath($\mathbf{p}_r$, $\mathbf{p}_i$)
    \If{feasible}
        \State $\mathbf{g} \gets$ Pose($\mathbf{p}_i$, YawToQuat($\psi_i$))
        \State NavigateTo($\mathbf{g}$)
        \State \Return $\mathbf{g}$
    \EndIf
\EndFor

\State \Return $\bot$
\end{algorithmic}
\end{algorithm}

\subsection{Approach module}
The approach module computes a safe approach pose suitable for VLA manipulation from a labeled object pose on a flat support surface (e.g., a tabletop marked as an obstacle in the occupancy grid). It isolates the surface region, recovers its boundary, estimates the surface normal with principal component analysis~\cite{WOLD198737}, and places the robot outside the edge with a user-defined offset while orienting the base perpendicular to the surface. Nav2~\cite{macenski2020marathon2} validates reachability and collision-free access; If the pose is infeasible, the module iterates over nearby edge candidates and reports failure only after exhausting valid options.

\section{VLA Model Fine-Tuning}
We fine-tuned the SmolVLA 450M model using an NVIDIA RTX 4090 GPU 16 GB VRAM. Fine-tuning was performed on a collected dataset comprising 50 pick-and-place episodes collected with SO-101 manipulator~\cite{cadene2024lerobot} by teleportation with a leader manipulator.

We used a batch size of 16 and a cosine decay scheduler with a learning rate of 0.0001 warmup~\cite{gotmare2018closerlookdeeplearning}. An AdamW optimizer~\cite{loshchilov2019decoupledweightdecayregularization} with a weight decay of 0.01 and warmup of 100 was used. Gradients were clipped to a norm of 10.0 to mitigate the explosion.

\section{HermesBot Mobile Manipulator}
\begin{figure}[t]
    \centering
    \includegraphics[width=\linewidth]{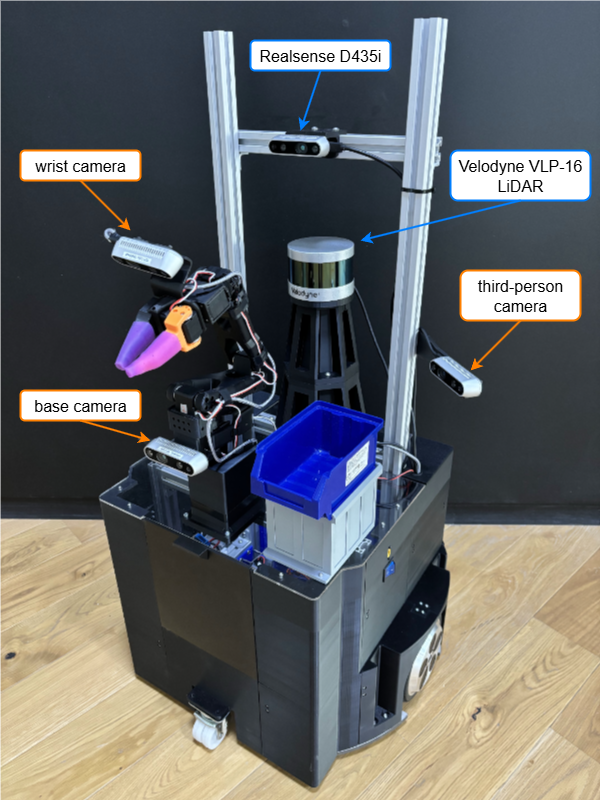}
    \caption{HermesBot mobile manipulator.}
    \label{fig:HermesBot}
\end{figure}

To support the deployment of AnywhereVLA, we developed a mobile manipulator platform tailored to integrated multi-modal sensing and computation. In Fig. \ref{fig:HermesBot}, The robot consists of a two-wheeled differential drive platform with the mounted SO-101 manipulator.

\subsection{Sensor Configuration}

The sensor suite is divided into navigational and VLA subsystems. Navigation is based on a Velodyne VLP-16 LiDAR and an Intel RealSense D435i RGB-D camera, integrated with \cite{zheng2024fastlivo2fastdirectlidarinertialvisual} for Visual-LiDAR-Inertial SLAM.

\begin{figure}[t]
\centering
\subfloat[Third-person-view camera]{\includegraphics[width=0.31\columnwidth]{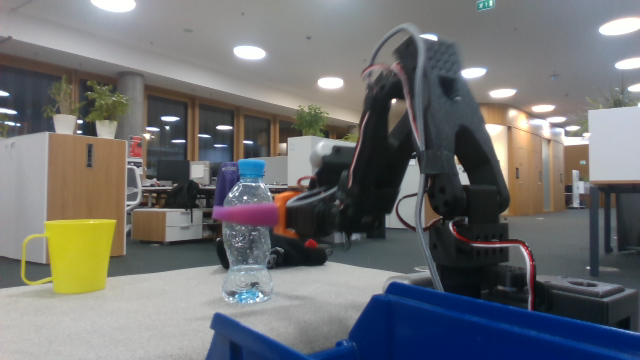}}\hfill
\subfloat[Wrist camera]{\includegraphics[width=0.31\columnwidth]{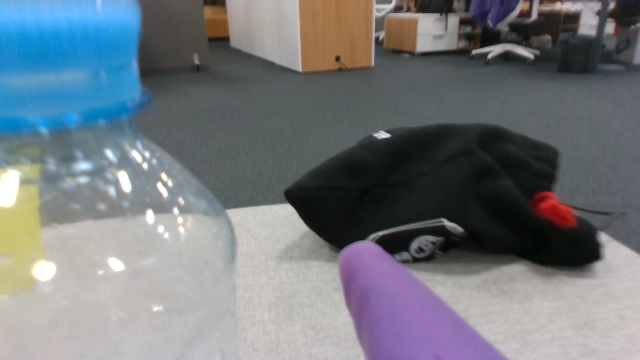}}\hfill
\subfloat[Base camera]{\includegraphics[width=0.31\columnwidth]{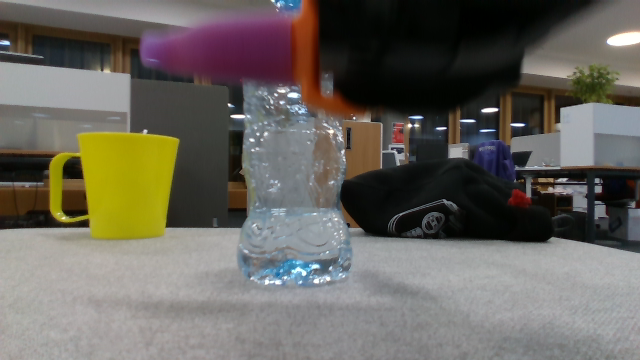}}
\caption{Intel Realsense D435 camera frames.}
\label{fig:vla}
\end{figure}

The VLA subsystem employs three Intel RealSense D435 cameras: a third-person-view camera angled to overview the manipulator and operating area (Fig. \ref{fig:vla}(a)), a wrist-mounted camera capturing the SO-101 manipulator's two-fingered gripper (Fig. \ref{fig:vla}(b)), and a base-mounted camera facing forward to monitor the manipulator's workspace (Fig. \ref{fig:vla}(c)).

\subsection{Computational Hardware}

The computational setup comprises an NVIDIA Jetson Orin NX 16Gb for GPU-accelerated tasks and an Intel NUC Core i7 32Gb for CPU-intensive operations. The Jetson Orin processes the perception-heavy SM and VLA Manipulation modules, while the NUC handles SLAM and navigation.

Table~\ref{tab:comp_dist} details the distribution of the pipeline modules, including inference frequencies and latencies measured during real-world deployment.

\begin{table}[h]
\centering
\small
\caption{Computational throughput of AnywhereVLA modules.}
\label{tab:comp_dist}
\resizebox{\columnwidth}{!}{%
\begin{tabular}{p{2.5cm} p{2.2cm} c c}
\toprule
Module & Computer & Frequency (Hz) $\uparrow$ & Process time (ms) $\downarrow$ \\
\midrule
SLAM & Intel NUC & 10 & 25 \\
Semantic Map & Jetson Orin & 15 & 45 \\
VLA & Jetson Orin & 15 & 20 \\
\bottomrule
\end{tabular}%
}
\end{table}

\begin{figure}[H]
    \centering
    \includegraphics[width=\linewidth]{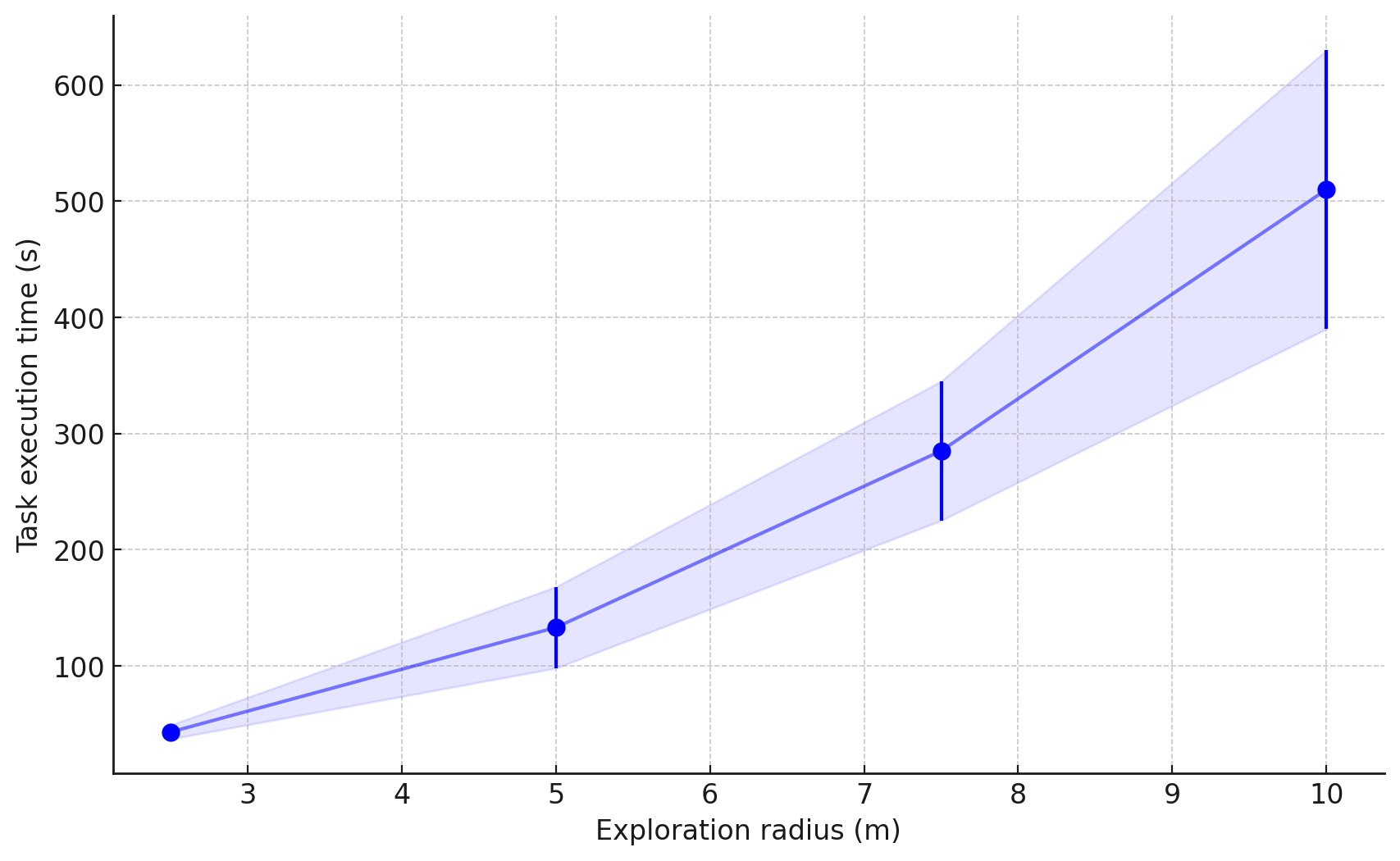}
    \caption{Total episode completion time.}
    \label{fig:time}
\end{figure}

\section{Experimental Results}
To evaluate the performance of AnywhereVLA architecture, we conducted comprehensive end-to-end experiments in diverse, unseen, indoor environments of university open space with dynamic clutter and human presence. We executed 50 pick-and-place episodes, where target objects were randomly placed within an exploration radius \(R_e\). In each episode, the robot was tasked with the command: \textbf{``Pick up the \textcolor{orange}{\texttt{<p>object</p>}} and place it in the \textcolor{blue}{\texttt{<p>area</p>}}. And bring the \textcolor{orange}{\texttt{<p>object</p>}} to \textcolor{red}{\texttt{<p>location</p>}}.''} In all 50 experiments we used a blue box on HermesBot mobile manipulator as \textbf{\textcolor{red}{\texttt{<p>location</p>}}}.
AnywhereVLA achieves an overall SR of 46\% with 85\% SR of VLA Manipulation module with fine-tuned SmolVLA compared to just 10\% overall SR without SmolVLA fine-tuning on our purpose-built pick-and-place dataset.

Due to the modular architecture of AnywhereVLA, in Table~\ref{tab:sr} we highlight SR of each component individually to get insights on their influence towards the overall performance. Errors in VLA manipulation were mainly due to the bottle slipping out of the gripper and AEE failed in 25\% cases being unable to find the desired object within radius $R_e$ due to cluttered tight spaces and corridors.

\begin{table}[t]
\centering
\caption{Success Rate of AnywhereVLA modules.}
\label{tab:sr}
\begin{tabular}{c|c}
\toprule
Module & SR (\%) $\uparrow$ \\
\midrule
SLAM & 100 \\
Active Environment Exploration & 75 \\
Navigation & 90 \\
Object Detection & 85 \\
VLA Manipulation & 80 \\
\bottomrule
\end{tabular}
\end{table}

To evaluate the feasibility of real-world deployment, we measured the total episode completion time of AnywhereVLA in a set of exploration radii \(R_e \in \{2.5, 5, 7.5, 10\}\)~m. In Fig.~\ref{fig:time}, each run of AnywhereVLA within 5m radius, which is equivalent to an average apartment, took under 133 seconds on average. And within 10m radius our architecture managed to successfully complete tasks under 10 minutes.

\section{Conclusion, Limitations and Future Work}
In this paper, we identify a shortcoming in the existing mobile manipulation architectures: the inability to explore open and large-scale indoor environments. To address this issue, we propose an AnywhereVLA, a modular framework for language-conditioned exploration and manipulation in large-scale novel indoor environments. The system combines classical SLAM, exploration, and navigation with a lightweight VLA fine-tuned for pick-and-place operations with SO-101 by TheRobotStudio. To evaluate our system, we conducted experiments in unseen dynamic university environments.
Deployed fully onboard with 3D Semantic Mapping and VLA on Jetson Orin NX and Active Environment Exploration and SLAM on Intel NUC, AnywhereVLA sustains real-time operation at $\geq 10$ Hz and achieves a $46\%$ overall success rate completing tasks under 2.5 minutes in $80m^2$ area.

While our language-conditioned exploration pipeline demonstrates robust performance in identifying and localizing target objects within unstructured environments, AnywhereVLA exhibits a notable limitation in enforcing precise spatial-semantic constraints specified in the natural language instruction. Specifically, our pipeline is unable to successfully comprehend cases such as \textbf{``Pick up the bottle \textcolor{red}{\texttt{from the table}} and place it in the blue box. And bring the bottle to me.''} Our architecture will explore to find the first bottle it detects, no matter where it is.

To address this limitation, future iterations of the pipeline could incorporate hierarchical semantic parsing and relational reasoning modules to disentangle object-level detection from spatial-prepositional constraints. One promising approach involves constructing a dynamic scene graph during exploration, where nodes represent detected objects and affordances, and edges encode probabilistic spatial relations derived from multimodal perception streams. Exploration policies could then be augmented with a graph-based reward signal that evaluates path efficiency and relational fidelity, penalizing deviations from the instructed configuration.. Alternatively, integrating a lightweight Vision-Language-Model for zero-shot relational query resolution (e.g., querying "is the bottle supported by the table?") could provide an end-to-end differentiable check prior to success attribution, thereby enhancing compositional generalization without substantial computational overhead. These enhancements would align the system more closely with human-like instruction following, particularly in cluttered, multi-object scenes prevalent in real-world robotic deployment.

\bibliographystyle{IEEEtran}
\bibliography{references}

\end{document}